\documentclass{article}

     \PassOptionsToPackage{sort&compress,authoryear}{natbib}
     
 \usepackage[final]{neurips_2023_ml4ps}

 \setcitestyle{authoryear,open={(},close={)}}

\usepackage[utf8]{inputenc} 
\usepackage[T1]{fontenc}    
\usepackage{url}            
\usepackage{booktabs}       
\usepackage{amsfonts}       
\usepackage{nicefrac}       
\usepackage{microtype}      
\usepackage[font=scriptsize]{caption}
\usepackage{xcolor}
\usepackage[many]{tcolorbox}
\usepackage{amssymb}
\usepackage{lipsum,graphicx,wrapfig}

\usepackage{graphicx,stackengine,scalerel}
\def\tang{\ThisStyle{\abovebaseline[0pt]{\scalebox{-1}{$\SavedStyle\perp$}}}}
\definecolor{generalcitation}{RGB}{96,170,236}
\definecolor{generalcite}{RGB}{0,0,175}
\usepackage[colorlinks=true,citecolor=generalcite,urlcolor=magenta]{hyperref}

\title{Attention-enhanced neural differential equations for physics-informed deep learning of ion transport}

\author{%
  Danyal Rehman \\ 
  {\small{Center for Computational Science and Engineering}} \\
  {\small{Massachusetts Institute of Technology (MIT)}}\\
  {\small{Cambridge, MA 02139, USA}} \\
\texttt{\href{mailto:drehman@mit.edu}{\textnormal{\color{generalcite}\small drehman@mit.edu}}} \\
  \And
  John H. Lienhard \\
  {\small{Department of Mechanical Engineering}} \\
  {\small{Massachusetts Institute of Technology (MIT)}}\\
  {\small{Cambridge, MA 02139, USA}} \\
\texttt{\href{mailto:lienhard@mit.edu}{\textnormal{\color{generalcite}\small lienhard@mit.edu}}} \\
}

\begin{document}
\setlength\parskip\baselineskip
\maketitle

\begin{abstract}\small
Species transport models typically combine partial differential equations (PDEs) with relations from hindered transport theory to quantify electromigrative, convective, and diffusive transport through complex nanoporous systems;\ however, these formulations are frequently substantial simplifications of the governing dynamics, leading to the poor generalization performance of PDE-based models. Given the growing interest in deep learning methods for the physical sciences, we develop a machine learning-based approach to characterize ion transport across nanoporous membranes. Our proposed framework centers around attention-enhanced neural differential equations that incorporate electroneutrality-based inductive biases to improve generalization performance relative to conventional PDE-based methods. In addition, we study the role of the attention mechanism in illuminating physically-meaningful ion-pairing relationships across diverse mixture compositions. Further, we investigate the importance of pre-training on simulated data from PDE-based models, as well as the performance benefits from hard vs.\ soft inductive biases. Our results indicate that physics-informed deep learning solutions can outperform their classical PDE-based counterparts and provide promising avenues for modelling complex transport phenomena across diverse applications.  
\end{abstract}\normalsize

\section*{Introduction and Background}
Modelling ion transport phenomena is a common problem observed across a host of applications that include ion-exchange through biological membranes \citep{gschwend2020discrete}, diffusing ionized gases in nuclear reactors \citep{PhysRevLett.81.3403}, and the transport of metal ions through polyamide nanopores \citep{ROY2015360,ritt2020ionization}. Across these applications, different models are typically used to describe the physics of the governing transport phenomena;\ however, the two most common approaches, derived from irreversible thermodynamics, are the Maxwell-Stefan formulations and the Nernst-Planck (NP) equations \citep{taylor1993multicomponent}. Maxwell-Stefan frameworks, although typically more accurate than the NP approach, can be used to model inter-species diffusion, yet require access to inter-species diffusion coefficients that become challenging to measure when large numbers of species are present \citep{krishna1997maxwell}. The NP equations, albeit simpler, introduce many assumptions and simplifications into the governing dynamics, which can adversely impact model performance making generalization a challenge \citep{REHMAN2023120325}. Given the advent of deep learning methods in the natural sciences, there are clear opportunities to address some of the shortcomings of classical PDE-based ion transport models \mbox{\citep{membranes11020128,rehman2023physicsconstrained}} through deep learning-based alternatives\footnote{Relevant research covering machine learning for PDEs and ion transport is detailed in Appendix \ref{sec:litreview}.}. \\ \\
In this work, we explore the use of attention-enhanced neural differential equations to model ion transport across polyamide nanopores \citep{chen2018neural}. We supplement classical neural differential equation models using the attention mechanism \citep{vaswani2017attention} and encode electroneutrality as an inductive bias \citep{rehman2023physicsconstrained} into the model architecture. Next, we pre-train the model on simulated data from PDE-based models supplemented with Gaussian noise to emulate experimental error, and then fine-tune the model on experimental data from over 750 measurements \citep{MICARI2020118117}. Further, we highlight the importance of the attention layers by illustrating their ability to learn physically-representative ion-pairing relationships across studied solutions \citep{AHDAB2020118072,AHDAB2021115037}. We also run ablations to ascertain the benefits of pre-training, while investigating the performance trade-offs between hard and soft inductive bias constraints. Lastly, we benchmark the performance of our approach relative to other competitive deep learning methods \citep{lecun1989handwritten,https://doi.org/10.48550/arxiv.1505.04597}. Using our proposed method, we show that it is possible to learn multi-species transport across nanoporous membranes and improve predictive performance relative to conventional PDE-based solutions \citep{GERALDES2008172}. 
\section*{Physics-informed Deep Learning Model}
\paragraph{Neural Ordinary Differential Equations}The hidden layer dynamics, $\textbf{h}(J_v)$, are parameterized by a first-order ordinary differential equation (ODE) that depends on transmembrane fluid flux, $J_v$:\
\begin{equation}
\frac{d{\textbf{h}(J_v)}}{dJ_v} = f_{\theta}(\textbf{h}(J_v), J_v; \theta)
\end{equation}
\begin{wrapfigure}{r}{0.39\textwidth}\vspace{-14px}
    \centering
    \includegraphics[width=0.423\textwidth]{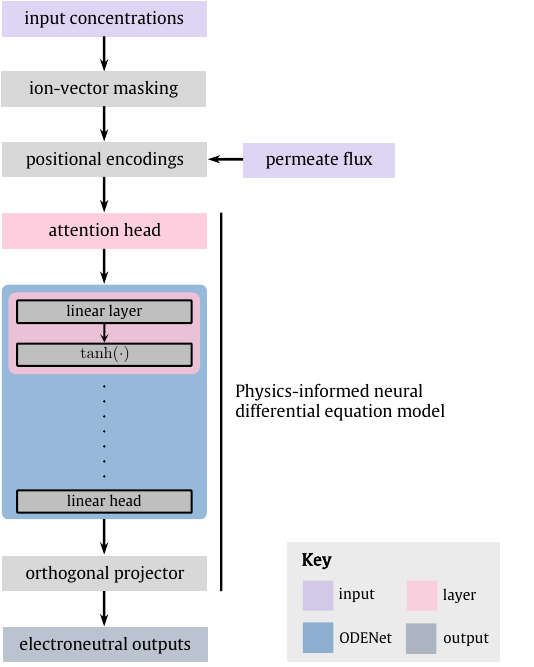}
    \caption{Physics-informed ODENet with an attention head and orthogonal projector for learning ionic context and predicting electroneutral outputs.}\vspace{-40px}
\end{wrapfigure}
\noindent with $J_v = \{0 \dots \mathcal{J}_v\}$, $\textbf{h} \in \mathbb{R}^d$, and $f_{\theta}\hspace{-2px}: [0, \mathcal{J}_v] \times \mathbb{R}^d \rightarrow \mathbb{R}^d$. Here, $d$ corresponds to the maximum number of charged solutes present across all datasets\footnote{We combine independent sets of experimental data all studying DuPont's FilmTec{\texttrademark} NF270 polyamide membrane. The set of ions present across solutions is $\mathcal{S}$\hspace*{2px}:\ \{Na$^+$, K$^+$, Li$^+$, Mg$^{2+}$, Ca$^{2+}$, Cl$^-$, SO$_4^{2-}$, NO$_3^-$\}.}. Since the distribution of ions varies across the studied datasets, we mask out absent ions prior to the positional encodings and attention layer.

The outputs of the neural differential equations correspond to scalar ion concentrations, $\textbf{h}(J_v)$, which are obtained by integrating over ODENet using the Tsitouras 5(4) numerical method \citep{simos2018fitted}. In ODENet, $\theta \in \Theta$, is a set of learnable parameters from some finite dimensional parameter space, $\Theta$ \citep{chen2018neural}. By learning the derivative of the output function, concentrations are uniformly Lipschitz continuous in $\textbf{h}(J_v)$ and continuous in $J_v$, enabling facile pre-training on classical PDE-based transport models \citep{https://doi.org/10.48550/arxiv.2202.02435,boral2023neural}. 

The model is comprised of five linear layers, each with ${\tanh}(\cdot)$ non-linearities applied to the outputs. Prior to the orthogonal projector, no point-wise activations are used. The network is trained using Adam with a batch size of 32 and an initial learning rate of $10^{-3}$ \citep{https://doi.org/10.48550/arxiv.1412.6980}. 

\paragraph{Attention Mechanism}
In language models, the attention mechanism serves as a means for learning semantic context \citep{vaswani2017attention};\ in the molecular or ionic setting, we can also leverage the attention mechanism to learn ionic context across diverse mixtures \citep{velickovic2018graph}. Using this approach, the model can identify governing ion-pairing relationships that dictate transport across polyamide nanopores \citep{REHMAN2023120325}. Attention is calculated as follows:\
\begin{equation}
\mathrm{Attention}(Q, K, V) = \mathrm{softmax}\left(\frac{QK^{\tang}}{\sqrt{d_k}}\right)V
\end{equation}
\noindent where $Q$, $K$, and $V$ are the query, key, and value matrices, obtained from $W_Q^{\tang} \in \mathbb{R}^{d_k}$, $W_K^{\tang} \in \mathbb{R}^{d_k}$, and $W_V^{\tang} \in \mathbb{R}^{d_k}$, respectively. In the reported work, we set $d_k = 8$ unless stated otherwise. 

\newpage\clearpage
\paragraph{Inductive Biases:\ Charge Conservation} 
For dissociated ions in fluid systems, electroneutrality is typically a conserved quantity in the bulk solution \citep{GUPTA2022115761,WANG202116906,REHMAN2023PHYSICSINFORMED}. The conservation law can be quantified as follows:\
\begin{equation}
\sum_{j = 1}^{d} z_j \textbf{h}_j(J_v) = 0, \hspace{10px} \forall J_v
\end{equation}
\noindent When treated as a hard constraint, we use the orthogonal projection of the hidden layer to ensure electroneutral outputs from the model. The projection is evaluated as follows:\ $z^{\tang}\textbf{h}_{\perp} = z^{\tang} \textbf{h} - z^{\tang} \textbf{h}_{\parallel}$, where $z \in \mathbb{R}^d$ corresponds to ion valences. During ablations, we elucidate the importance of the inductive bias by also applying it as a soft constraint. In this case, the electroneutrality term is simply appended to the loss functions as a regularization constraint to be minimized \citep{RAISSI2019686}. 
\paragraph{Training Regime and Augmentations}
We first pre-train the neural solver on simulated data from the Donnan--Steric Pore Model with Dielectric Exclusion (DSPM--DE):\ a well-established PDE-based approach that involves solving the Nernst-Planck equations \citep{GERALDES2008172}. Details and derivation of the PDE-based model and regression formulation are provided in prior work \citep{WANG2021118809,REHMAN2022100034}. The pre-training loss is expressed as follows:\
\begin{equation}
\mathcal{L}^{\mathrm{PDE}}(\textbf{h}, \textbf{h}^{\mathrm{PDE}}) = \frac{1}{kd}\sum_{i = 1}^k\sum_{j = 1}^d \left[\textbf{h}_{j}(J_{v, i}) - {\textbf{h}^{\mathrm{PDE}}_{j}}(J_{v, i})\right]^2
\end{equation}
\noindent Subsequently, we freeze the first three layers of the network and fine-tune the remaining two using measurement data fitted with Gaussian statistics to emulate experimental uncertainty:\ 
\begin{equation}
\mathcal{L}^{\mathrm{exp}}(\textbf{h}, \textbf{h}^{\mathrm{exp}}) = \frac{1}{nd}\sum_{i = 1}^n\sum_{j = 1}^d \left[\textbf{h}_{j}(J_{v, i}) - {\textbf{h}^{\mathrm{exp}}_{j}}(J_{v, i})\right]^2, \hspace{10px}{\textbf{h}_{j}^{\mathrm{exp}}}(J_{v, i}) \sim \mathcal{N}(\mu_{ij}, \sigma_{ij}^2) \ \forall i,j
\end{equation} 
\noindent where $n$ corresponds to the number of flux measurements taken per species. 

\section*{Results and Discussion}
\begin{wrapfigure}{r}{0.6\textwidth}
    \centering\hspace*{-7px}
    \includegraphics[width=0.61\textwidth]{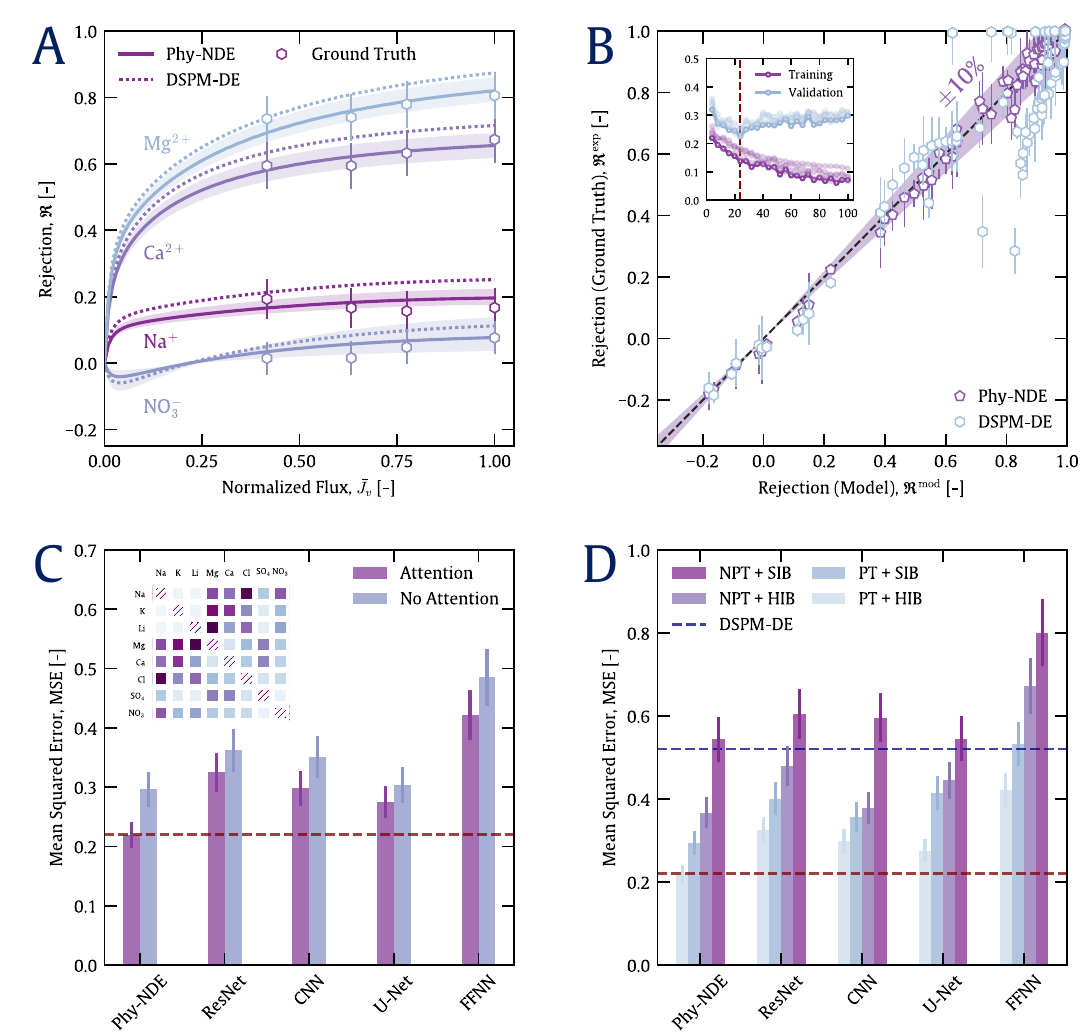}
    \caption{\textbf{(A)}:\ Ion rejection predictions as a function of flux using the physics-informed ODENet and PDE-based DSPM--DE. \textbf{(B)}:\ Parity plot illustrating general predictive performance across the test set using both the proposed ODENet and DSPM--DE. \textbf{(C)}:\ The MSE loss achieved across deep learning methods with and without the attention mechanism included. The inset corresponds to a sample learned attention matrix for a given ionic composition from the test set. \textbf{(D)}:\ The implications of pre-training and hard vs.\ soft inductive biases on predictive performance across a set of assessed deep learning methods. Key:\ \textbf{PT}:\ pre-trained with simulated data;\ \textbf{NPT}:\ no pre-training with simulated data;\ \textbf{HIB}:\ hard inductive bias;\ \textbf{SIB}:\ soft inductive bias.}\vspace{-20px}
    \label{fig:resultsanddiscussion}
\end{wrapfigure}

\paragraph{Predictive Performance}For a sample ionic composition from the test set, we predict ion rejection rollouts as a function of flux:\ $\mathfrak{R}_j^{\mathrm{mod}}(J_v) \triangleq \left(1 - \left[{\textbf{h}_j(J_v)}/{c_{j, \mathrm{in}}}\right]\right)$. In Fig.\ \ref{fig:resultsanddiscussion}\textbf{A}), we observe that the neural model outperforms the classical PDE-based method for a given rollout, while we note generally superior performance across the full test set in Fig.\ \ref{fig:resultsanddiscussion}\textbf{B}). $\pm$10\% confidence bounds are included to illustrate the strong agreement achieved by the neural approach relative to the PDE-based model, which shows substantial deviations from the ground truth across a large number of test samples. 
\paragraph{Implications of Attention}In Fig.\ \ref{fig:resultsanddiscussion}\textbf{C}), we benchmark the accuracy of our physics-informed ODENet relative to other deep learning methods and quantify the performance benefits of the attention mechanism. Our approach outperforms other ML-based approaches with the U-Nets achieving the closest MSE. This is likely due to the smooth profiles generated by ODENet for unseen fluxes that closely mirror experimental observation;\ other methods are unable to capture this continuity leading to inferior performance on the test data. For all conducted tests, we maintained a similar number of model parameters across benchmarks to ensure a fair comparison. In all cases, we note that the inclusion of the attention layer improves predictive performance. \\ \\
Further, in the inset, we include a sample of the attention matrix with all ions reported. We note that the attention mechanism clearly learns the importance of valence and ionic size in transport, as seen by the elevated scores present for ions with the largest differences in radius and charge. Even more interestingly, we note that the importance of preserving electroneutrality is also learned:\ in cases of negative rejection $-$ as demonstrated by NO$_3^-$ in Fig.\ \ref{fig:resultsanddiscussion}\textbf{A}) $-$ the transport of the partner cation(s) or anion(s) is accelerated to conserve charge;\ the attention matrix is able to accurately identify the pertinent ion-pairs instrumental in achieving electroneutrality. In the example shown, we see that the attention given to both Cl$^-$ and NO$_3^-$ by Na$^+$ is high;\ this makes physical sense as SO$_4^{2-}$ is too large and immobile to be transported through the polyamide nanopores meaning that Cl$^-$ and NO$_3^-$ are the primary ions carried across to achieve electroneutrality \citep{AHDAB2021117425}. This is similarly observed in Fig.\ \ref{fig:resultsanddiscussion}\textbf{A}), where the presence of three cations means that NO$_3^-$ transport must be expedited to ensure electroneutral outputs (this is exemplified by the negative rejection observed in NO$_3^-$). These findings clearly illustrate the value of the attention mechanism in learning and characterizing ion transport across polyamide nanopores.
\vspace*{-15px}
\paragraph{Pre-training and Inductive Biases}In Fig.\ \ref{fig:resultsanddiscussion}\textbf{D}), we quantify the impact of pre-training on simulated data from mechanistic models, and integrating inductive biases as hard constraints rather than soft regularization terms. By not pre-training on classical PDE-based models, we note that the resultant MSE is nearly 40\% higher than when it is included. Despite the shortcomings of these PDE-based models \citep{WANG2021118809}, using them to improve the quality of the intermediate embeddings through pre-training substantially improves predictive performance on the downstream task. 

In addition, our results demonstrate that pre-training on simulated data is even more impactful than treating the inductive biases as hard constraints. This is likely because in the data-limited regime, the model struggles to learn meaningful representations of the governing transport phenomena without appropriate guidance from the PDE-based methods;\ as a result, the inductive biases are insufficient in providing substantive signal to learn physically-representative trends. In the extreme case, when pre-training is not performed, and inductive biases are treated as soft constraints, we observe that predictive performance is even worse than the PDE-derived baseline across all deep learning-based methods investigated. 



\section*{Conclusion}
In this work, we propose attention-based neural differential equations for learning multi-ionic transport across nanoporous membranes. The model employs attention layers to learn physically-representative ion-pairing relationships that govern transport phenomena. We illustrate this through reported attention matrices that elucidate the role of ion valence in species transport. In addition, we run ablation studies to investigate the importance of pre-training the neural approach on classical PDE-based models. In data-limited settings, like the regime we are operating in, learning high quality embeddings through pre-training on classical models is imperative to achieving competitive performance against non deep learning alternatives. We also investigate the performance implications of treating charge conservation inductive biases as hard and soft constraints. We show that including them in either capacity outperforms classical PDE-based models across most deep learning methods (barring simple feed forward neural networks). Our results also highlight that transitioning from soft constraints to hard constraints drops MSE loss on the test set by 10-20\%. These findings speak to the potential of neural methods to serve as robust alternatives to PDE-based models that often struggle to meet our performance requirements across diverse input and operating conditions.

\newpage\clearpage
\begin{ack}
The authors thank the Centers for Mechanical Engineering Research and Education at MIT and SUSTech (MechERE Centers at MIT and SUSTech) for partially funding this research. D.R.\ acknowledges financial support provided by a fellowship from the Abdul Latif Jameel World Water and Food Systems (J-WAFS) Lab and fellowship support from the Martin Family Society of Fellows.
\end{ack}

\footnotesize
\bibliography{neurips_ML_manuscript}

\newpage\clearpage
\appendix
\section{Related Work}
\label{sec:litreview}\normalsize
\paragraph{Deep Learning for PDEs}
Some recent scientific machine learning research has considered the development of neural operators, like the Fourier Neural Operator (FNO) \citep{li2021fourier} and DeepONet \citep{Lu_2021}, for applications to thermal-fluid sciences \citep{li2023geometryinformed}, carbon capture \citep{wen2023real}, and even quantum systems \citep{zhang2023artificial}. These methods typically map function spaces to function spaces to learn grid-independent dynamics of diverse PDEs. Other work has focused on finite-dimensional neural alternatives to numerical methods like neural ordinary differential equations \citep{chen2018neural}, universal differential equations \citep{rackauckas2021universal}, physics-informed neural networks \citep{RAISSI2019686}, and Clifford group equivariant networks \citep{brandstetter2023clifford}, to name a few. Ongoing efforts have also tried to make these neural methods more accurate and/or data-efficient through the integration of symmetry-derived inductive biases \citep{pmlr-v162-brandstetter22a,mialon2023selfsupervised}, while others have focused on improving long-term prediction accuracies \citep{wan2023evolve,lippe2023pderefiner}. For ion transport specifically, there have been some efforts to leverage neural methods to supplement mechanistic models for fouling prediction \citep{DEJAEGHER2021118028} and membrane monomer design \citep{zhang2020deep}, but the development of new deep learning-based process models remains a field of ongoing research.  

\paragraph{Mechanistic Ion Transport Models}
The first ion transport models across nanoporous membranes were derived from irreversible thermodynamics in the 1960s \citep{kedem1963permeability}. Since then, substantial improvements have been proposed, with both the Maxwell-Stefan frameworks \citep{KRAAIJEVELD1995163}, and the extended Nernst-Planck equations-based Donnan--Steric Pore Model with Dielectric Exclusion (DSPM--DE) \citep{BOWEN20021121}. Other recent models are typically  extensions and variants of these models that have been further developed in recent years \citep{wang2021salt,wang2023significance}. Despite this progress, these models typically struggle to generalize to new compositions due to the introduction of highly-constraining model simplifications and assumptions \citep{REHMAN2023120325}. These simplifications are most often introduced as closure models to estimate parameters under nano-confinement;\ however, these functional relationships typically overconstrain the model, preventing accurate predictions across diverse operating conditions \citep{WANG2021118809}. 

\end{document}